\documentclass[preprint, authoryear]{elsarticle}

\usepackage{bbm,latexsym,setspace,graphicx,natbib,amsmath,subfigure,multirow,multicol,fullpage,rotating,comment,color,subfig}

\usepackage{amssymb}

\usepackage[T1]{fontenc}

\usepackage{url}

\newcommand{\E}{\mbox{E}}

\usepackage{tikz}
\usetikzlibrary{plotmarks,quotes,angles,positioning,decorations.pathreplacing,shapes.misc,automata,arrows,positioning,calc}
\usetikzlibrary{decorations.pathreplacing,shapes.geometric}

\tikzstyle{startstop} = [rectangle, rounded corners, minimum width=3cm, minimum height=1cm,text centered, text width=3cm, draw=black,fill=Forestpurple!30]
\tikzstyle{process} = [rectangle, minimum width=3cm, minimum height=1cm, text centered, text width=3cm, draw=black, fill=MidnightBlue!30]
\tikzstyle{problem} = [ellipse, minimum width=1cm, minimum height=0.5cm, text centered,text width=2.5cm, draw=black,  fill=WildStrawberry!45]
\tikzstyle{lack} = [chamfered rectangle, chamfered rectangle xsep=0.5cm, minimum width=1cm, minimum height=1cm, text centered, draw=red, thick,text width=1cm]
\tikzstyle{pub} = [ellipse, minimum width=1cm, minimum height=1cm, text centered, draw=Forestpurple, line width=1mm,text width=1cm]
\tikzstyle{arrow} = [thick,->,>=stealth]

\tikzstyle{prior} = [rectangle, rounded corners, minimum width=3cm, minimum height=1cm,text centered, draw=black]
\tikzstyle{covariates} = [ellipse, minimum width=1cm, minimum height=1cm, text centered, draw=black]
\tikzstyle{model} = [rectangle, minimum width=2cm, minimum height=1cm, text centered, draw=black]

\tikzstyle{prior2} = [rectangle, rounded corners, minimum width=0.5cm, minimum height=0.5cm,text centered, draw=black]

\definecolor{jade}{rgb}{0.0, 0.66, 0.42}

\begin{document}

\title{Synthesizing Property \& Casualty Ratemaking Datasets using Generative Adversarial Networks}

\author[1]{Marie-Pier C\^ot\'e} 
\author[2,corr]{Brian Hartman} 
\author[1]{Olivier Mercier}
\author[2]{Joshua Meyers}
\author[2]{Jared Cummings} 
\author[2]{Elijah Harmon}

\address[1]{\'Ecole d'actuariat, Universit\'e Laval, Qu\'ebec, QC, Canada}
\address[2]{Department of Statistics, Brigham Young University, Provo, UT, USA}
\address[corr]{Corresponding Author: hartman@stat.byu.edu}

\begin{abstract}
Due to confidentiality issues, it can be difficult to access or share interesting datasets for methodological development in actuarial science, or other fields where personal data are important. We show how to design three different types of generative adversarial networks (GANs) that can build a synthetic insurance dataset from a confidential original dataset. The goal is to obtain synthetic data that no longer contains sensitive information but still has the same structure as the original dataset and retains the multivariate relationships. In order to adequately model the specific characteristics of insurance data, we use GAN architectures adapted for multi-categorical data: a Wassertein GAN with gradient penalty (MC-WGAN-GP), a conditional tabular GAN (CTGAN) and a Mixed Numerical and Categorical Differentially Private GAN (MNCDP-GAN). For transparency, the approaches are illustrated using a public dataset, the French motor third party liability data. We compare the three different GANs on various aspects: ability to reproduce the original data structure and predictive models, privacy, and ease of use. We find that the MC-WGAN-GP synthesizes the best data, the CTGAN is the easiest to use, and the MNCDP-GAN guarantees differential privacy.
\end{abstract}

\maketitle

\section{Introduction}

In order to improve the quality and accuracy of the models used in insurance practice, methodological developments must be tested on the type of data they are meant to model. Unfortunately, insurance claims data at the individual policyholder or claimant level are highly confidential. Just like medical records, these data cannot be publicly shared unless meaningful covariates are erased. This lack of publicly available data slows down the methodological developments in actuarial science.

Take as an example loss reserving. With the improved availability of computing resources, reserving methods that traditionally use aggregate information may now model individual claims. To this end, \citet{Antonio/Plat:2014}, \citet{Pigeon/etal:2013, Pigeon/etal:2014}, and \citet{Wuthrich:2018ml} proposed micro-level reserving models and illustrated their efficacy on confidential datasets. Because the data is confidential, it is difficult to compare the methods to each other or to new methods yet to be developed. Additionally, the research is not easily reproducible, even when the code is shared.

The lack of publicly available data was discussed in \cite{Gabrielli/Wuthrich:2018}, where the authors provide an \textsf{R} program for simulating insurance claim development patterns. A Gaussian copula with appropriate margins generates the features, and the different parts of the development process are modeled with successive neural nets. The simulation machine accommodates only a few covariates; the generation of a large number of features with the Gaussian copula could lead to unrealistic combinations of factor levels. In this paper, we propose to synthesize insurance data with a generative adversarial network (GAN).

A GAN is a deep learning model that was introduced in \cite{Goodfellow/etal:2014}. It consists of two competing neural networks: one that generates fake data, the so-called generator, and a second, the discriminator, that is trained to identify whether the data is real or fake. During the training process, the generator adapts in order to fool the discriminator, which means that it learns to generate fake data that is indistinguishable from the real data. The resulting GAN could thus be used to simulate a synthetic dataset, that is completely fake, but still has the structure of real data.

\citet{Frid/etal:2018} use GANs to generate synthetic data in order to augment a small imaging dataset and improve the performance of the classification of liver lesion. As explained in \cite{Papernot/etal:2017}, a method based on GANs can provide strong privacy for sensitive training data. \citet{Choi/etal:2017} proposed the medGAN architecture to synthesize realistic patient records. Their motivation is similar to ours: patient records are highly confidential but extremely valuable for developing new models and statistical methods. The structure of patient record data is also closer to that of insurance data than most of the deep learning literature, focusing on unstructured data such as images. Images (and pixels) are continuous, whereas most of claimant characteristics are categorical variables, which adds complexity as one cannot interpolate between discrete classes to create fake records. \cite{Camino/etal:2018} adapted the medGAN and the Wassertein GAN with gradient penalty (WGAN-GP) of \cite{Gulrajani/etal:2017} for multi-categorical variables. 

Deep learning has gained interest in recent actuarial research. \cite{Schelldorfer/Wuthrich:2019} analyze the French motor third party liability claims dataset studied in Section~\ref{sec:CS} with a generalized linear model embedded in a neural network. \cite{Wuthrich:2018} use neural networks for chain-ladder reserving. To our knowledge however, only \citet{kuo2019generative} has used one type of GAN in actuarial applications so far.

In this paper, we introduce other GANs to the actuarial science literature and adapt the metrics to be appropriate for Poisson count data. Although there exists publicly available frequency datasets to develop and test pricing methods, they are really toy datasets compared to those that are kept confidential, due to the low number of policyholder or other covariates, like telematics or spatial information. We present and test three architectures. The first one, in Section~\ref{sec:method}, is based on the multi-categorical adaptation by \cite{Camino/etal:2018} of the WGAN-GP. Section~\ref{sec:CTGAN} presents the conditional tabular GAN of \cite{xu2019modeling}, applied to ratemaking data in \citet{kuo2019generative}. The last model is what we call the MNCDP-GAN, detailed in Section~\ref{sec:MNCDP}, which is an adaptation of the differentially private GAN with autoencoder developed in \cite{Tantipongpipat/etal:2019}. The MNCDP-GAN is the only one that incorporates differential privacy, which is the gold standard for guaranteeing that data can be shared without confidentiality issues. Section~\ref{sec:CS} shows a case study where the three architectures are tested on the French MTPL dataset,  publicly available in the \textsf{R} package \texttt{CASdatasets} \citep{CASData}. All of the code is available in the GitHub repository for this paper.\footnote{\url{https://github.com/brianmhartman/Anonymizing-Ratemaking-Datasets-using-GANs}} Section~\ref{sec:conclu} concludes the paper and is followed by \ref{app:B} and \ref{app:A} which detail the setup and tuning of the multi-categorical and continuous WGAN-GP and the MNCDP-GAN, respectively.


\section{Multi-categorical Wassertein GAN}
\label{sec:method}

Let us first introduce the general framework of generative adversarial networks. The training of a GAN is a game between two competing networks: the generator and the discriminator. The generator $G$ is a neural net with parameter vector $\theta_g$ that takes in argument a vector of random noise $Z$ with distribution $F_z$, and maps it to the space of the data we wish to model. Usually, the components of the vector $Z$ are independent standard Gaussian random variables, and the dimension of $Z$ is lower than the that of the data. The resulting $G(Z;\theta_g)$ is a fake data point, and its distribution is denoted by $F_g$. 

The goal of the training procedure is therefore to find a good approximation $F_g$ of the unknown distribution of a true data point~$X$, denoted $F_x$. To achieve this goal, a competing network, the discriminator $D$ with parameter vector $\theta_d$, learns to determine whether a data point is real or fake. To this end, the parameters $\theta_d$ of $D$ are trained to maximize the expected score of a real data point $\E_X\{D(X;\theta_d)\}$ and to minimize the expected score of a synthetic data point $\E_Z[D\{G(Z;\theta_g);\theta_d\}]$. To achieve the goal of generating realistic data points, the parameters $\theta_g$ of the generator are trained to maximize the discriminator's score on a \emph{fake} data point $\E_Z[D\{G(Z;\theta_g);\theta_d\}]$. Combining the two problems together, the two networks aim to solve
$$
\min_{\theta_g}\max_{\theta_d} \E_X[\log\{D(X;\theta_d)\}] + \E_Z\left[\log[1-D\{G(Z;\theta_g);\theta_d\}]\right].
$$
This optimization problem amounts to minimizing the Jensen-Shannon divergence between $F_x$ and~$F_g$. In practice, this leads to serious convergence issues, partly solved by training $D$ and $G$ in turn with minibatches. 

\begin{figure}
    \centering
\begin{tikzpicture}[node distance=3.25cm]
\node (noise) [text width=1.5cm, text centered] {Noise $Z$ \\ $\sim F_z$\\ $\mathcal{N}(\mathbf{0},\mathbf{I}_k)$};
\node (gen) [rectangle, minimum width=1.6cm, minimum height=3cm, fill=orange!30,  text width=1.5cm, text centered, right of = noise, xshift=-0.75cm] {Generator $G(Z; \theta_g)$};
\node (fake) [right of = gen, text width=2.5cm, text centered, xshift=-0.25cm] {Fake group of\\ policyholders $\sim F_g$};
\node (real) [above of = fake, text width=2.5cm, text centered] {Real group of\\ policyholders $\sim F_X$};
\node (critic) [rectangle, minimum width=1.5cm, minimum height=3cm, fill=blue!30,  text width=1.5cm, text centered, above right of = fake, yshift=-0.63cm, xshift=-0.15cm] {Critic $D(\cdot)$\\ \footnotesize{$\mathcal{D}$: set of 1-Lipschitz functions}};
\node (data) [rectangle, text centered, draw=black, above of = gen] {Data $X$};

\node (decreal) [rectangle,rounded corners,text centered, draw=black, fill=gray!30, right of = critic, xshift=-1.2cm,text width=1.2cm, text centered] {Score of realism };

\draw [->, thick] (noise)--(gen);
\draw [->, thick] (gen)--(fake);
\draw [->, thick] (data)--(real);

\draw [->, thick] (fake)--(critic);
\draw [->, thick] (real)--(critic);

\draw [->, thick] (critic)--(decreal);

\node (backprop) [right of=decreal, xshift=-2cm] {\textbf{Loss}};
\path (backprop) edge[bend right=90, ->, color=blue] node [above right, color=blue, xshift=-2.2cm, yshift=0.1cm, align=right] {$\max_{D\in\mathcal{D}} \E\{D(X)\}
-\E[D\{G(Z;\theta_g)\}]$
} (critic);
\draw [->, orange] (backprop) [out=-90, in=-40] to (gen) node[below, xshift = 3.3cm, yshift=-0.1cm] {$\displaystyle{\max_{\theta_g} \E[D\{G(Z;\theta_g)\}]}$};

\end{tikzpicture}
    \caption{WGAN Schema. The arrows represent the flow of the training process.}
    \label{fig:WGAN-schema}
\end{figure}
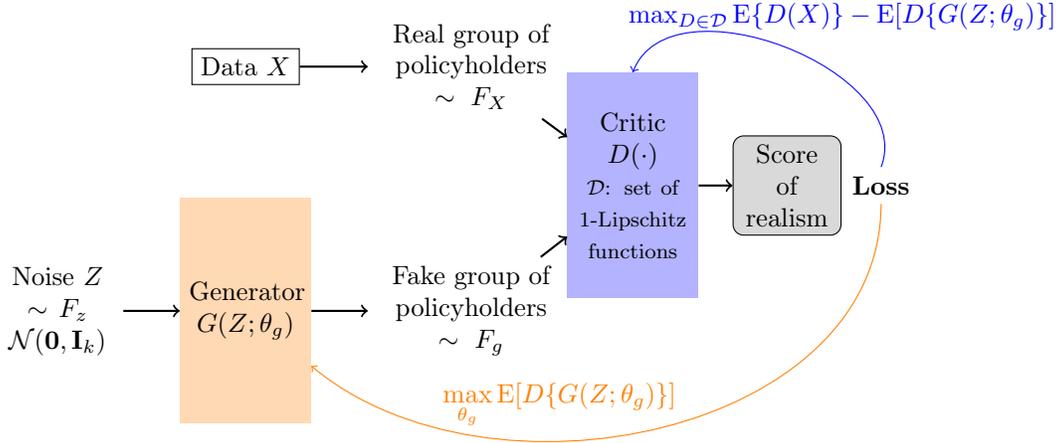

To solve some of the convergence issues, \cite{Arjovsky/etal:2017} advocate the use of the Wassertein-1 distance between $F_x$ and $F_g$, that is, they consider the problem
$$
\min_{\theta_g}\max_{D\in \mathcal{D}} \E_X\{D(X)\} - \E_Z\left[D\{G(Z;\theta_g)\}\right],
$$
where $ \mathcal{D}$ is the set of 1-Lipschitz functions. This change in the objective function leads to the Wassertein generative adversarial network, or WGAN. The discriminator in a WGAN is called the \emph{critic}, as it is real valued rather than a binary classifier. The WGAN is depicted schematically in Figure~\ref{fig:WGAN-schema} for policyholder claim data $X$. The black arrows represent the forward flow of information in the network, while the colored arrows represent the flow of the training process for the generator (orange) and the critic (blue).

Some tactics are needed to enforce the Lipschitz constraints on $D$. In this regard, the gradient penalty (GP) developed by \cite{Gulrajani/etal:2017} greatly improves the training of the WGAN. In their WGAN-GP, the authors take advantage of the fact that a differentiable Lipschitz function has gradients with norm at most 1 everywhere. A tuning parameter $\lambda>0$ is introduced, and the objective of the WGAN-GP is
$$
\min_{\theta_g}\max_{\theta_d} \E_X\{D(X;\theta_d)\} - \E_Z\left[D\{G(Z;\theta_g);\theta_d\}\right]+\lambda \E_{\hat{X}}[\{||\nabla_{\hat{x}}D(\hat{X};\theta_d)||_2-1\}^2],
$$
where $\hat{X} \stackrel{d}{=} U X+(1-U)G(Z;\theta_g)$, and $U$ is uniformly distributed on the interval $(0,1)$, so that the distribution $F_{\hat{x}}$ of $\hat{X}$ is obtained by sampling uniformly along lines between pairs of points sampled from $F_x$ and $F_g$. For details on the motivation, the reader is referred to \cite{Gulrajani/etal:2017}.

In practice, if $m\in \mathbb{N}$ is the size of the minibatch with observations $x_1,\ldots,x_m$, random noise vectors $z_1,\ldots,z_m$ and independent uniform samples $u_1,\ldots,u_m$, then we let $\hat{x}_i=u_ix_i+(1-u_i)G(z_i;\theta_g)$ and the discriminator loss is approximated by
$$
\mathcal{L}_d = \frac{1}{m}\sum_{i=1}^m -D(x_i,\theta_d)+D\{G(z_i;\theta_g);\theta_d\}+\lambda\{||\nabla_{\hat{x}_i}D(\hat{x}_i;\theta_d)||_2-1\}^2
$$
while the generator loss is simply
$$
\mathcal{L}_d = \frac{1}{m}\sum_{i=1}^m -D\{G(z_i;\theta_g);\theta_d\}.
$$
Note that higher values of the critic $D$ indicate fake samples.

The WGAN and the WGAN-GP were developed in the context of image generation tasks. In the current application however, we wish to synthesize tabular insurance data, in which some variables are categorical with multiple levels. An application closer to ours was considered by \cite{Camino/etal:2018}, where the target data contains many multi-categorical variables. \cite{Camino/etal:2018} modified the generator of the WGAN-GP so that, after the model output, there is a dense layer in parallel for each categorical variable followed by a softmax activation function. Then, the results are concatenated to yield the final generator output.

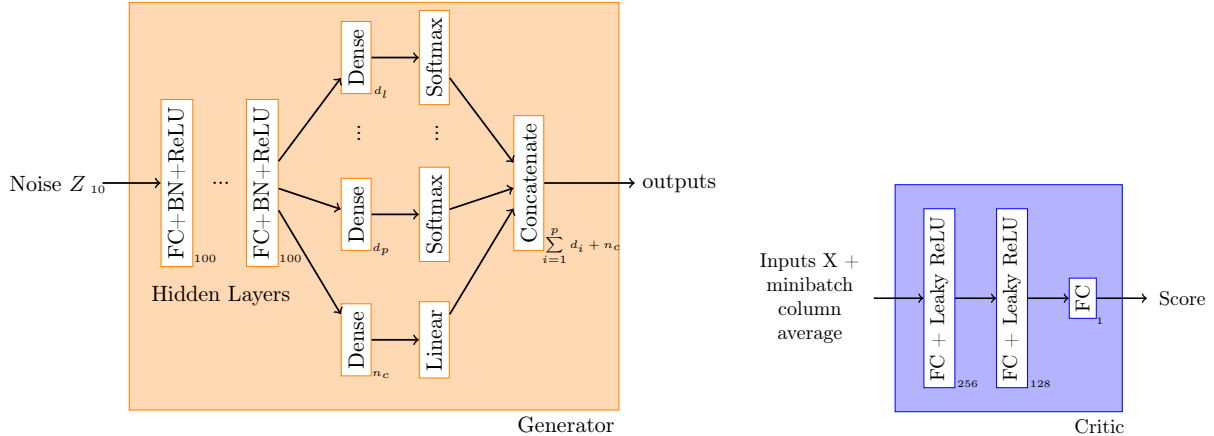
\begin{figure}
    \centering
    \resizebox{.6\textwidth}{!}{
    \begin{tikzpicture}[node distance=1 cm]
\node (noise) [text width=1.5cm, text centered] {Noise $Z$};
\node (rect) [rectangle, minimum width=7.8cm, minimum height=6.5cm, draw=orange, fill=orange!30, text width=2cm, text centered, right of = noise, xshift=4.2cm,yshift=-0.37cm] {};
\node (FC1) [draw=orange, fill=white!100, right of= noise,rotate=90,anchor=north, yshift=-0.8cm] {FC+BN+ReLU};
\begin{tiny}
\node (dim1) [right of= FC1,below of=FC1, xshift=-0.55cm, yshift=-0.2cm] {100};
\end{tiny}
\begin{tiny}
\node (dim7) [right of= noise,below of=noise, xshift=-0.2cm, yshift=0.9cm, ] {10};
\end{tiny}
\node (FC2) [right of= FC1, xshift=-0.3cm] {...}; 
\node (FC3) [draw=orange, fill=white!100, right of= FC2,rotate=90,anchor=north, yshift=0.6cm] {FC+BN+ReLU}; 
\begin{tiny}
\node (dim2) [right of= FC3,below of=FC3, xshift=-0.55cm, yshift=-0.2cm] {100};
\end{tiny}
\node (MC1) [draw=orange, fill=white!100, right of= FC3,rotate=90,anchor=north, xshift=2cm,yshift=-0.25cm] {Dense}; 
\begin{tiny}
\node (dim3) [right of= MC1,below of=MC1, xshift=-0.6cm, yshift=0.45cm, ] {$d_l$};
\end{tiny}
\node (MC1act) [draw=orange, fill=white!100, right of= MC1,rotate=90,anchor=north] {Softmax}; 
\node  [below of= MC1,rotate=90,anchor=north, yshift=0.1cm, xshift=-0.2cm] {...};
\node  [below of= MC1act,rotate=90,anchor=north, yshift=0.1cm, xshift=-0.2cm] {...};
\node (underb) [right of=FC1, yshift=-1.8cm, xshift=-0.3cm]{Hidden Layers};
\node (MC2) [draw=orange, fill=white!100, right of= FC3,rotate=90,anchor=north, xshift=-0.5cm,yshift=-0.25cm] {Dense};
\begin{tiny}
\node (dim4) [right of= MC2,below of=MC2, xshift=-0.6cm, yshift=0.45cm, ] {$d_p$};
\end{tiny} 
\node (MC2act) [draw=orange, fill=white!100, right of= MC2,rotate=90,anchor=north] {Softmax};
\node (MC3) [draw=orange, fill=white!100, right of= FC3,rotate=90,anchor=north, xshift=-2.5cm,yshift=-0.25cm] {Dense};
\begin{tiny}
\node (dim.1) [right of= MC3,below of=MC3, xshift=-0.6cm, yshift=0.45cm, ] {$n_c$};
\end{tiny} 
\node (MC3act) [draw=orange, fill=white!100, right of= MC3,rotate=90,anchor=north] {Linear};  
\node (name) [right of=MC3act, yshift=-1.35cm, xshift=1.1cm] {Generator};
\node (concat) [draw=orange, fill=white!100, right of= FC3,rotate=90,anchor=north, yshift=-3cm] {Concatenate}; 
\begin{tiny}
\node (dim6) [right of= concat,below of=FC1, xshift=5.45cm, yshift=0cm, ] {$\displaystyle{\sum^p_{i=1} d_i + n_c}$};
\end{tiny}
\node  (out) [right of= concat, xshift=1.4cm] {outputs}; 
\draw [->, thick] (noise)--(FC1);
\draw [->, thick] (FC3)--(MC1);
\draw [->, thick] (MC1)--(MC1act);
\draw [->, thick] (MC1act)--(concat);
\draw [->, thick] (FC3)--(MC2);
\draw [->, thick] (MC2)--(MC2act);
\draw [->, thick] (MC2act)--(concat);
\draw [->, thick] (FC3)--(MC3);
\draw [->, thick] (MC3)--(MC3act);
\draw [->, thick] (MC3act)--(concat);
\draw [->, thick] (concat)--(out);
\end{tikzpicture} %
}
\resizebox{.39\textwidth}{!}{
\begin{tikzpicture}[node distance=0.5 cm] %
\node  (Input) [fill=white!100, yshift=0cm, xshift=0cm, text width=2cm, text centered] {Inputs X + minibatch column average};
\node (rect) [rectangle, minimum width=4cm, minimum height=4cm, draw=blue, fill=blue!30, text width=2cm, text centered, right of = Input,
 xshift=3cm] {};
\node  (item1) [right of=Input, draw=blue, fill=white!100, rotate=90,anchor=north, yshift=-1.5cm, xshift=0cm] {FC + Leaky ReLU};
\node  (item2) [draw=blue, right of=item1, fill=white!100, rotate=90,anchor=north, yshift=-0.5cm, xshift=0cm] {FC + Leaky ReLU};
\node  (item3) [right of=item2, fill=white!100, yshift=0cm, xshift=2.5cm, text width=1cm, text centered] {Score};
\node (FC) [draw=blue, fill=white!100, right of= item2,rotate=90,anchor=north, yshift=-0.5cm] {FC};
\begin{tiny}
\node (dim1) [right of= FC,below of=FC, xshift=-0.2cm, yshift=0.1cm, ] {1};
\end{tiny}
\begin{tiny}
\node (dim2) [right of= item1,below of=item1, xshift=0cm, yshift=-1cm, ] {256};
\end{tiny}
\begin{tiny}
\node (dim3) [right of= item2,below of=item2, xshift=0cm, yshift=-1cm, ] {128};
\end{tiny}
\node (titre) [above of=Input, yshift=-2.75cm, xshift=5.1cm]{Critic};
\draw [->, thick] (Input)--(item1);
\draw [->, thick] (FC)--(item3);
\draw [->, thick] (item2)--(FC);
\draw [->, thick] (item1)--(item2);
\end{tikzpicture}
}
    \caption{Architecture inside the generator (orange) and the critic (blue) for our multi-categorical and continuous WGAN. The dimensions $d_1,\ldots,d_p$ represent the number of levels in categorical variables $1,\ldots, p$. FC stands for fully connected and BN stands for Batch normalization.}
    \label{fig:MC-WGAN}
\end{figure}

As in Camino et al. (2018), our generator's architecture has one dense layer with dimension matching the number of levels for each multi-categorical variable. We also add one dense layer with linear activation and dimension $n_c$ which is equal to the number of continuous variables. The architecture of the generator and the critic in our multi-categorical and continuous WGAN-GP, or MC-WGAN-GP, is depicted in Figure~\ref{fig:MC-WGAN}. Further details about hyperparameter optimization are available in \ref{app:B}.

\section{CTGAN}
\label{sec:CTGAN}

Another possible path to simulating insurance claim data is called a conditional tabular GAN or CTGAN \citep{xu2019modeling}. This method was applied to ratemaking data in \citet{kuo2019generative}. Additionally, Kuo developed an \textsf{R} wrapper for this software to make it easily accessible to insurance practitioners more familiar with \textsf{R} than Python. Starting from his code, we slightly adjusted the preamble to improve the application consistency on our machines and we adjusted the preprocessing slightly, but other than that the overall code remained the same. Our version of the code is available in the GitHub repository for this paper. 

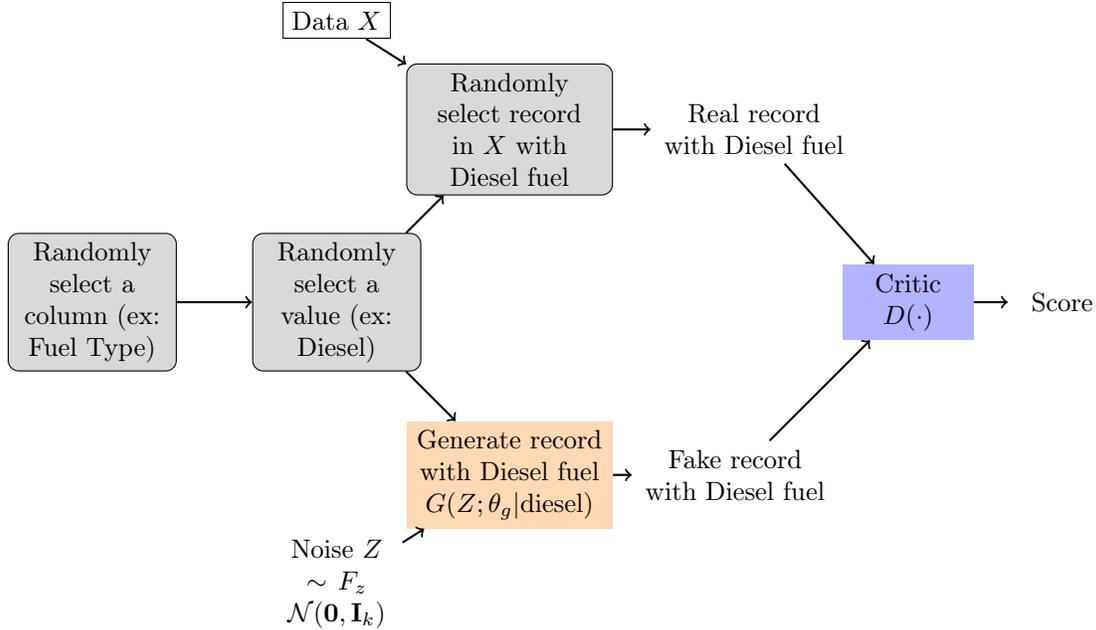
\begin{figure}
    \centering
\begin{tikzpicture}[node distance=3.25cm]

\node (randval) [rectangle,rounded corners,text centered, draw=black, fill=gray!30, text width=2cm, text centered] {Randomly select a value (ex: Diesel)};

\node (randcol) [rectangle,rounded corners,text centered, draw=black, fill=gray!30, text width=2cm, text centered, left of = randval] {Randomly select a column (ex: Fuel Type)};

\node (noise) [text width=1.5cm, text centered, below of =randval, yshift=-0.5cm] {Noise $Z$ \\ $\sim F_z$\\ $\mathcal{N}(\mathbf{0},\mathbf{I}_k)$};

\node (gen) [rectangle, minimum width=1.6cm, minimum height=1cm, fill=orange!30,  text width=2.5cm, text centered, below right of = randval] {Generate record with Diesel fuel $G(Z; \theta_g| \mbox{diesel})$};

\node (fake) [right of = gen, text width=2.5cm, text centered, xshift=-0.25cm] {Fake record with Diesel fuel};

\node (findreal) [rectangle,rounded corners,text centered, draw=black, fill=gray!30, above right of = randval, text width=2.5cm, text centered] {Randomly select record in $X$ with Diesel fuel};

\node (real) [right of = findreal, text width=2.5cm, text centered] {Real record with Diesel fuel};

\node (critic) [rectangle, minimum width=1.5cm, minimum height=1cm, fill=blue!30,  text width=1.5cm, text centered, above right of = fake] {Critic $D(\cdot)$};

\node (data) [rectangle, text centered, draw=black, above of = randval, yshift=0.5cm] {Data $X$};

\node (decreal) [right of = critic, xshift=-1.2cm,text width=1.2cm, text centered] {Score};

\draw [->, thick] (randcol)--(randval);
\draw [->, thick] (randval)--(gen);
\draw [->, thick] (randval)--(findreal);
\draw [->, thick] (findreal)--(real);

\draw [->, thick] (noise)--(gen);
\draw [->, thick] (gen)--(fake);
\draw [->, thick] (data)--(findreal);

\draw [->, thick] (fake)--(critic);
\draw [->, thick] (real)--(critic);

\draw [->, thick] (critic)--(decreal);

\end{tikzpicture}
    \caption{CTGAN Schema. The model flow is illustrated for the case when the selected column is the Fuel Type and the selected value for that column is Diesel.}
    \label{fig:CTGAN-schema}
\end{figure}

The CTGAN simulates records one by one. It first randomly selects one of the variables (say fuel type, diesel or gasoline). Then, it randomly selects a value for that variable (say diesel). Following \cite{kuo2019generative}, we use the true data frequency to sample the value rather than the log-frequency as suggested in \cite{xu2019modeling}. Given that value for that variable, the algorithm finds a matching row from the training data (in this example, it randomly selects a true observation with a diesel-powered car). It also generates the rest of the variables conditioning on it being diesel-powered. The generated and true rows are sent to the critic which gives a score. Figure \ref{fig:CTGAN-schema} summarizes the CTGAN procedure.

\begin{figure}
    \centering
    \resizebox{.6\textwidth}{!}{
    \begin{tikzpicture}[node distance=1 cm]
    
\node (in) [text width=1.5cm, text centered] {Inputs};

\node (rect) [rectangle, minimum width=8.2cm, minimum height=5.7cm, draw=orange, fill=orange!30, text width=2cm, text centered, below of = in, xshift=2.3cm,yshift=-2.3cm] {};

\node (noise) [below of = in, draw=black, fill=white!100, text width=1.5cm, text centered, yshift = -0.5cm] {Noise $Z$ and \emph{cond}};

\node (FC1) [draw=orange, fill=white!100, below of= noise, yshift=-0.8cm] {FC+BN+ReLU};
\begin{tiny}
\node (dim1) [right of= FC1,below of=FC1, xshift=0.05cm, yshift=0.6cm] {256};
\end{tiny}

\node (FC3) [draw=orange, fill=white!100, below of= FC1, yshift=-0.8cm] {FC+BN+ReLU}; 
\begin{tiny}
\node (dim2) [right of= FC3,below of=FC3, xshift=0.05cm, yshift=0.6cm] {256};
\end{tiny}
\node (MC1) [draw=orange, fill=white!100, above right of= FC1,rotate=90,anchor=north, xshift=1cm, yshift=-1.5cm] {Dense}; 
\begin{tiny}
\node (dim3) [right of= MC1,below of=MC1, xshift=-0.6cm, yshift=0.45cm] {$d_l$};
\end{tiny}
\node (MC1act) [draw=orange, fill=white!100, right of= MC1,rotate=90,anchor=north] {Gumbel}; 
\node  [below of= MC1,rotate=90,anchor=north, yshift=0.1cm, xshift=-0.7cm] {...};
\node  [below of= MC1act,rotate=90,anchor=north, yshift=0.1cm, xshift=-0.7cm] {...};

\node (MC2) [draw=orange, fill=white!100, below right of= FC1,rotate=90,anchor=north, xshift=-1cm, yshift=-1.5cm] {Dense};
\begin{tiny}
\node (dim4) [right of= MC2,below of=MC2, xshift=-0.6cm, yshift=0.45cm, ] {$d_p$};
\end{tiny} 
\node (MC2act) [draw=orange, fill=white!100, right of= MC2,rotate=90,anchor=north] {Gumbel};
 
\node (name) [below of=FC3, yshift=-0.3cm, xshift=5.6cm] {Generator};

\node (concat) [draw=orange, fill=white!100, right of= FC1,rotate=90,anchor=north, yshift=-4cm] {Concatenate}; 
\begin{tiny}
\node (dim6) [right of= concat,below of=FC1, xshift=4.75cm, yshift=0cm, ] {$\displaystyle{\sum^p_{i=1} d_i}$};
\end{tiny}
\node  (out) [right of= concat, xshift=1.4cm] {outputs}; 
\draw [->, thick] (in)--(noise);
\draw [->, thick] (noise)--(FC1);
\draw [->, thick] (FC1)--(FC3);
\draw [->, thick] (noise.south west) to [out=230,in=120] (FC3.north west);

\draw [->, thick] (noise)--(MC1);
\draw [->, thick] (FC1.north east)--(MC1);
\draw [->, thick] (FC3.north east)--(MC1);

\draw [->, thick] (MC1)--(MC1act);
\draw [->, thick] (MC1act)--(concat);

\draw [->, thick] (noise.south east)--(MC2);
\draw [->, thick] (FC1.south east)--(MC2);
\draw [->, thick] (FC3)--(MC2);

\draw [->, thick] (MC2)--(MC2act);
\draw [->, thick] (MC2act)--(concat);
\draw [->, thick] (concat)--(out);
\end{tikzpicture} %
}
\resizebox{.39\textwidth}{!}{
\begin{tikzpicture}[node distance=0.5 cm] %
\node  (Input) [fill=white!100, yshift=0cm, xshift=0cm, text width=1.5cm, text centered] {10 fake and 10 real records};
\node (rect) [rectangle, minimum width=4cm, minimum height=5cm, draw=blue, fill=blue!30, text width=2cm, text centered, right of = Input,
 xshift=3cm] {};
\node  (item1) [right of=Input, draw=blue, fill=white!100, rotate=90,anchor=north, yshift=-1.5cm, xshift=0cm] {FC + Leaky ReLU + drop};
\node  (item2) [draw=blue, right of=item1, fill=white!100, rotate=90,anchor=north, yshift=-0.5cm, xshift=0cm] {FC + Leaky ReLU + drop};
\node  (item3) [right of=item2, fill=white!100, yshift=0cm, xshift=2.5cm, text width=1cm, text centered] {Score};
\node (FC) [draw=blue, fill=white!100, right of= item2,rotate=90,anchor=north, yshift=-0.5cm] {FC};
\begin{tiny}
\node (dim1) [right of= FC,below of=FC, xshift=-0.2cm, yshift=0.1cm, ] {1};
\end{tiny}
\begin{tiny}
\node (dim2) [right of= item1,below of=item1, xshift=0cm, yshift=-1.6cm, ] {256};
\end{tiny}
\begin{tiny}
\node (dim3) [right of= item2,below of=item2, xshift=0cm, yshift=-1.6cm, ] {256};
\end{tiny}
\node (titre) [above of=Input, yshift=-3.25cm, xshift=5.1cm]{Critic};
\draw [->, thick] (Input)--(item1);
\draw [->, thick] (FC)--(item3);
\draw [->, thick] (item2)--(FC);
\draw [->, thick] (item1)--(item2);
\end{tikzpicture}
}
    \caption{Architecture inside the generator (orange) and the critic (blue) for the CTGAN. The dimensions $d_1,\ldots,d_p$ represent the number of levels in categorical variables $1,\ldots, p$. All variables are assumed categorical. The input of the generator is Gaussian random noise and the condition \emph{cond} of the feature value that was randomly selected (see Figure~\ref{fig:CTGAN-schema}). FC stands for fully connected, BN for Batch normalization, Gumbel for the Gumbel softmax activation, and drop for dropout.}
    \label{fig:CTGAN-gen-critic}
\end{figure}
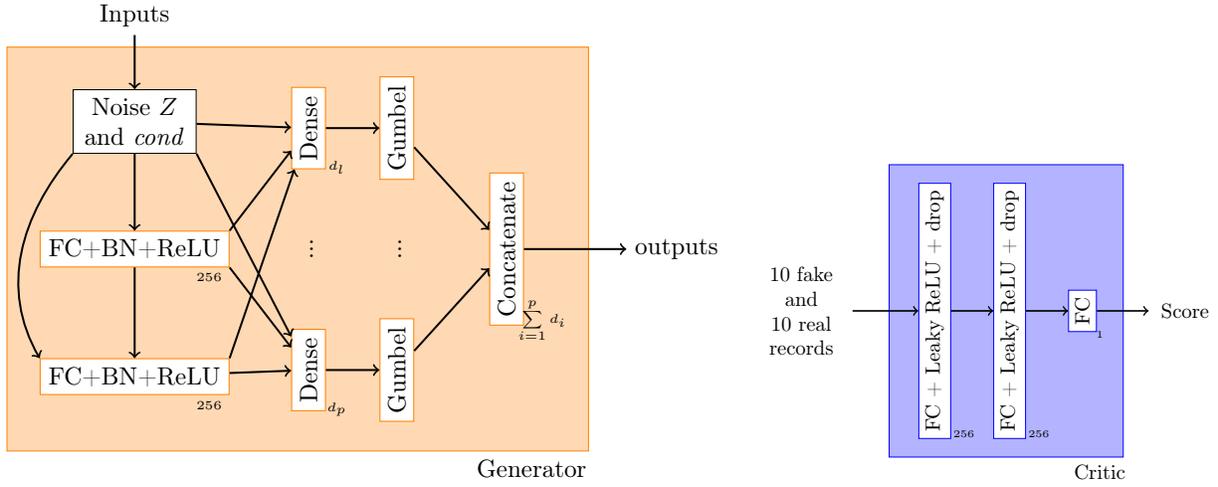

 Figure~\ref{fig:CTGAN-gen-critic} zooms inside the architecture of the generator and the critic. Both the critic (blue) and the generator (orange) use two fully-connected layers to attempt to capture all relationships between the columns. The generator uses skip connections and allows only for categorical variables; see \cite{xu2019modeling} for the non-trivial extension with continuous variables. An additional sophistication of the CTGAN is the use of the PacGAN framework \citep{Lin/etal:2018} in the discriminator, where ten samples are provided in each pac to prevent the mode collapse issue.

Like the previously discussed MC-WGAN-GP, the CTGAN does not incorporate privacy protections, though that could be possible to develop, as hinted in \citet{kuo2019generative}. This implementation also only simulates entirely discrete data, which is a disadvantage when working with continuous information such as exposure. 

\section{MNCDP-GAN}
\label{sec:MNCDP}

The Mixed Numerical and Categorical Differentially Private GAN (MNCDP-GAN) tries to solve the drawbacks of the two other GANs. It includes an autoencoder and a WGAN. The main advantage of this architecture, introduced in \cite{Tantipongpipat/etal:2019}, is that the generator works in a latent space of encoded variables, which can be easier to model adequately than the original structured data. The training can be done in a differentially private (DP) manner, allowing a DP guarantee on the generated dataset.

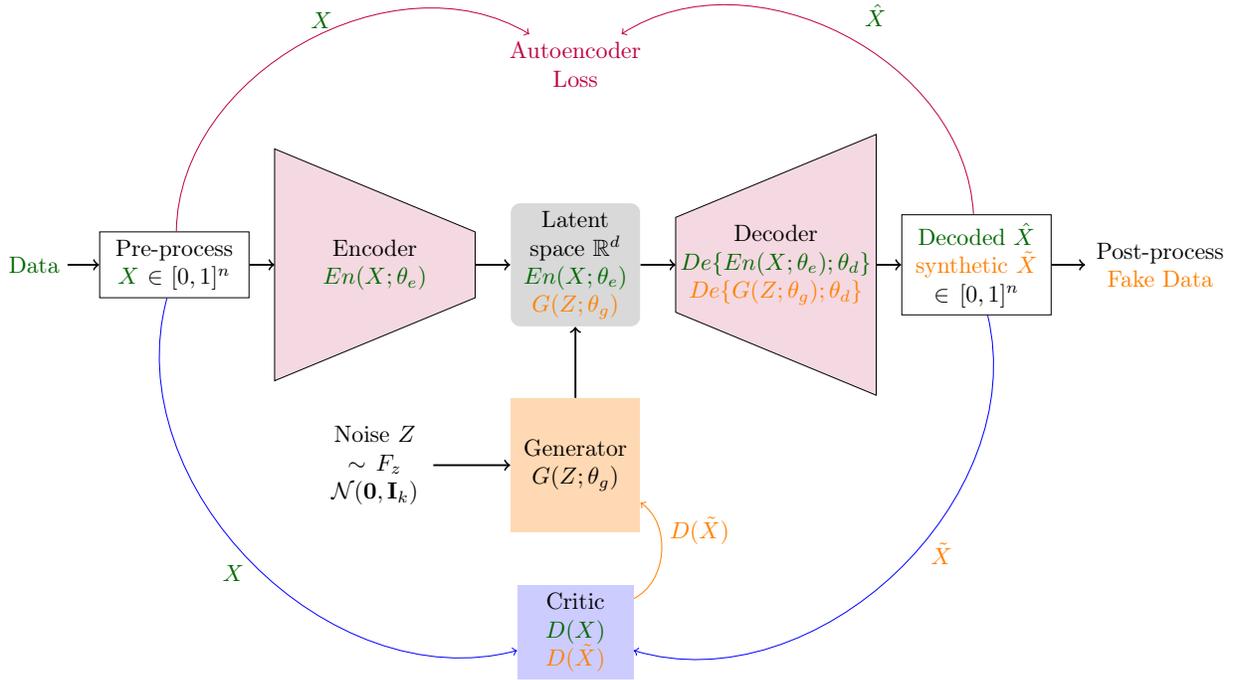
\begin{figure}
    \centering
 \resizebox{\textwidth}{!}{
    \begin{tikzpicture}[node distance=3cm]

\node (noise) [text width=1.5cm, text centered] {Noise $Z$ \\ $\sim F_z$\\ $\mathcal{N}(\mathbf{0},\mathbf{I}_k)$};

\node (gen) [rectangle, minimum width=1.6cm, minimum height=2cm, fill=orange!30,  text width=1.7cm, text centered, right of = noise, xshift=0cm] {Generator $G(Z; \theta_g)$};

\node (latent) [rectangle,rounded corners, minimum width=1.6cm, minimum height=1cm, fill=gray!30,  text width=1.7cm, text centered, above of = gen, xshift=0cm] {Latent space $\mathbb{R}^d$\\ \textcolor{green!40!black}{$En(X;\theta_{e})$}\\ \textcolor{orange}{$G(Z;\theta_g)$}};

\node (Enc) [trapezium, trapezium angle=67.5, draw,inner xsep=0pt,outer sep=0pt, minimum height=2cm,shape border rotate=270, fill=purple!15,  text width=3cm, text centered, left of = latent, xshift=0cm] {Encoder\\ \textcolor{green!40!black}{$En(X;\theta_{e})$}};

\node (pre) [rectangle, text centered, draw=black, left of = Enc, text width=2cm] {Pre-process\\ $\textcolor{green!40!black}{X}\in[0,1]^n$};

\node (data) [rectangle, text centered, left of = pre, xshift=0.9cm] {\textcolor{green!40!black}{Data}};

\node (Dec) [trapezium, trapezium angle=67.5, draw,inner xsep=0pt,outer sep=0pt, minimum height=2cm,shape border rotate=90, fill=purple!15,  text width=3cm, text centered, right of = latent, xshift=0cm] {Decoder \textcolor{green!40!black}{$De\{En(X;\theta_{e});\theta_{d}\}$}\\ \textcolor{orange}{$De\{G(Z;\theta_g);\theta_{d}\}$}};

\node (post) [rectangle, text centered, draw=black, right of = Dec, text width=2cm] {\textcolor{green!40!black}{Decoded $\hat{X}$} \textcolor{orange}{synthetic $\tilde{X}$}\\ $\in[0,1]^n$};

\node (auto) [rectangle, text centered, color=purple, above of = latent, text width=2.3cm] {Autoencoder Loss};

\node (fake) [right of = post, text width=2cm, text centered, xshift=-0.25cm] {Post-process\\ \textcolor{orange}{Fake Data}};

\node (critic) [rectangle, minimum width=1.5cm, minimum height=1cm, fill=blue!20,  text width=1.5cm, text centered, below of = gen, yshift=0.5cm, xshift=0cm] {Critic \textcolor{green!40!black}{$D(X)$} \\ \textcolor{orange}{$D(\tilde{X})$}};

\draw [->, thick] (noise)--(gen);
\draw [->, thick] (data)--(pre);
\draw [->, thick] (pre)--(Enc);
\draw [->, thick] (Enc)--(latent);
\draw [->, thick] (latent)--(Dec);
\draw [->, thick] (Dec)--(post);
\draw [->, thick] (post)--(fake);

\draw [->, thick] (gen)--(latent);

\path (pre) edge[bend right=60, ->, color=blue] node [below left, xshift=0cm, yshift=0.1cm, align=right, color=green!40!black] {$X$} (critic);
\path (post) edge[bend left=60, ->, color=blue] node [above right, color=orange, xshift=0.4cm, yshift=0cm, align=right] {$\tilde{X}$} (critic);

\path (pre) edge[bend left=60, ->, color=purple] node [above right, xshift=0cm, yshift=0.1cm, align=right,color=green!40!black] {$X$} (auto);
\path (post) edge[bend right=60, ->, color=purple] node [above right, xshift=0.2cm, yshift=0cm, align=right, color=green!40!black] {$\hat{X}$} (auto);

\path (critic) edge[bend right=60, ->, color=orange] node [above right, xshift=0cm, yshift=0cm, align=right, color=orange] {$D(\tilde{X})$} (gen);

\end{tikzpicture}}
    \caption{MNCDP-GAN Schema. The orange relates to the generated data while the green relates to the original data. The colored arrows represent the flow into the loss for the training of each network; autoencoder in red, generator in orange and discriminator in blue. For DP training, noise is added in training the decoder and the critic.}
    \label{fig:MNCDP-schema}
\end{figure}

As depicted in Figure~\ref{fig:MNCDP-schema}, the original data is first pre-processed (one-hot encodings for categorical variables and either binning or min-max standardization for continuous variables), resulting in vectors defined in $[0,1]^n$ that are fed into an encoder shrinking the dimension to $d<n$, a hyper-parameter. Then, a decoder takes in entry the encoded variable in the latent space $\mathbb{R}^d$ and outputs data in the format $[0,1]^n$, which is subsequently post-processed to lead to data in the original format. This architecture is called an autoencoder and is used in many neural network applications. In our context, the autoencoder creates the latent space in dimension~$d$, which is easier to learn for the generator as it has less structure than the original data space. The generator takes in random noise and outputs a vector in the latent space, which can then be decoded by the decoder to produce a synthesized record. The critic is trained with the Wassertein loss and compares the generated data before postprocessing with the preprocessed original data. 

In Figure~\ref{fig:MNCDP-schema}, the autoencoder flow and training are depicted in red, the flow of the data in the autoencoder and the critic is indicated in green, and the flow of the generated data through the decoder and the critic is highlighted in orange font. It is assumed, and reasonable, that the postprocessing step can be done using public knowledge and does not affect the DP quality of the model. The DP training is done by injecting noise in the decoder and the critic, for more details, refer to \cite{Tantipongpipat/etal:2019}. 

The level of differential privacy that is achieved by the model (including both the autoencoder and the GAN) is quantified by the value $\epsilon>0$. This value relates to how different an analysis may be if one datapoint is added or removed. If the dataset $X'$ is identical to $X$ except for one added data point, then an $(\epsilon,\delta)$ differentially private analysis $\mathcal{M}$ satisfies
$$
\Pr\{\mathcal{M}(X)\in S\} \leq e^{\epsilon}\Pr\{\mathcal{M}(X')\in S\}+\delta
$$
for $\delta>0$, any such $X,X'$ and $S\subseteq Range(\mathcal{M})$ \citep[see, e.g.,][]{Dwork/Roth:2014}. A smaller $\epsilon$ represents stronger privacy guarantees, but as more noise is added to the training, it also comes with decreased performance. The values of $\epsilon$ and $\delta$ for our procedure are obtained through a \emph{privacy accountant} as explained in \cite{Tantipongpipat/etal:2019}. Further details on our implementation are available in~\ref{app:A}.

\section{Case Study}
\label{sec:CS}

To show the value of the three approaches in a reproducible manner and to compare their effectiveness in producing synthetic data, we use a well-known publicly available dataset for the case study. It contains a set of 412,748 French motor third-party liability policies observed in a single year \citep{CASData}. There are eight explanatory variables in the data:
 \begin{itemize}
     \item Exposure: the number of car-years on the policy, bounded between 0 and 1 (we removed the few records with exposure greater than one)
     \item Power: an ordered categorical variable which describes the power of the vehicle
     \item CarAge: the vehicle age in years
     \item DriverAge: the primary driver age in years
     \item Brand: the vehicle brand divided in the following groups: A -- Renault, Nissan, and Citroen, B -- Volkswagen, Audi, Skoda, and Seat, C -- Opel, General Motors, and Ford, D -- Fiat, E -- Mercedes, Chrysler, and BMW, F-- Japanese (except Nissan) and Korean, G -- other.
     \item Gas: diesel or regular
     \item Region: the policy region in France
     \item Density: number of inhabitants per km$^2$ in the home city of the driver
 \end{itemize}

Brand, Gas, Power, and Region are all categorical variables and the other four are numeric (continuous or discrete). Since the CTGAN only accommodates categorical variables, the continuous variables are binned for that setup. For the MNCDP models we will show four different DP levels:

\begin{itemize}
    \item $\epsilon = \infty$ is labeled on the plots as "MNCDPInfty": no differential privacy
    \item $\epsilon = 100,000$ labeled as "MNCDP100k"
    \item $\epsilon = 10,000$ labeled as "MNCDP10k"
    \item $\epsilon = 5$ is labeled as "MNCDP5": strong differential privacy
\end{itemize}

We simulate a dataset of the same size as the original dataset using each of the GANs, and we compare the univariate distributions in the generated samples with the univariate distributions in the real data. If the methods are able to faithfully reproduce the original data, then we expect the distributions to be similar.

\begin{figure}[b]
\centering
    \includegraphics[width=\textwidth]{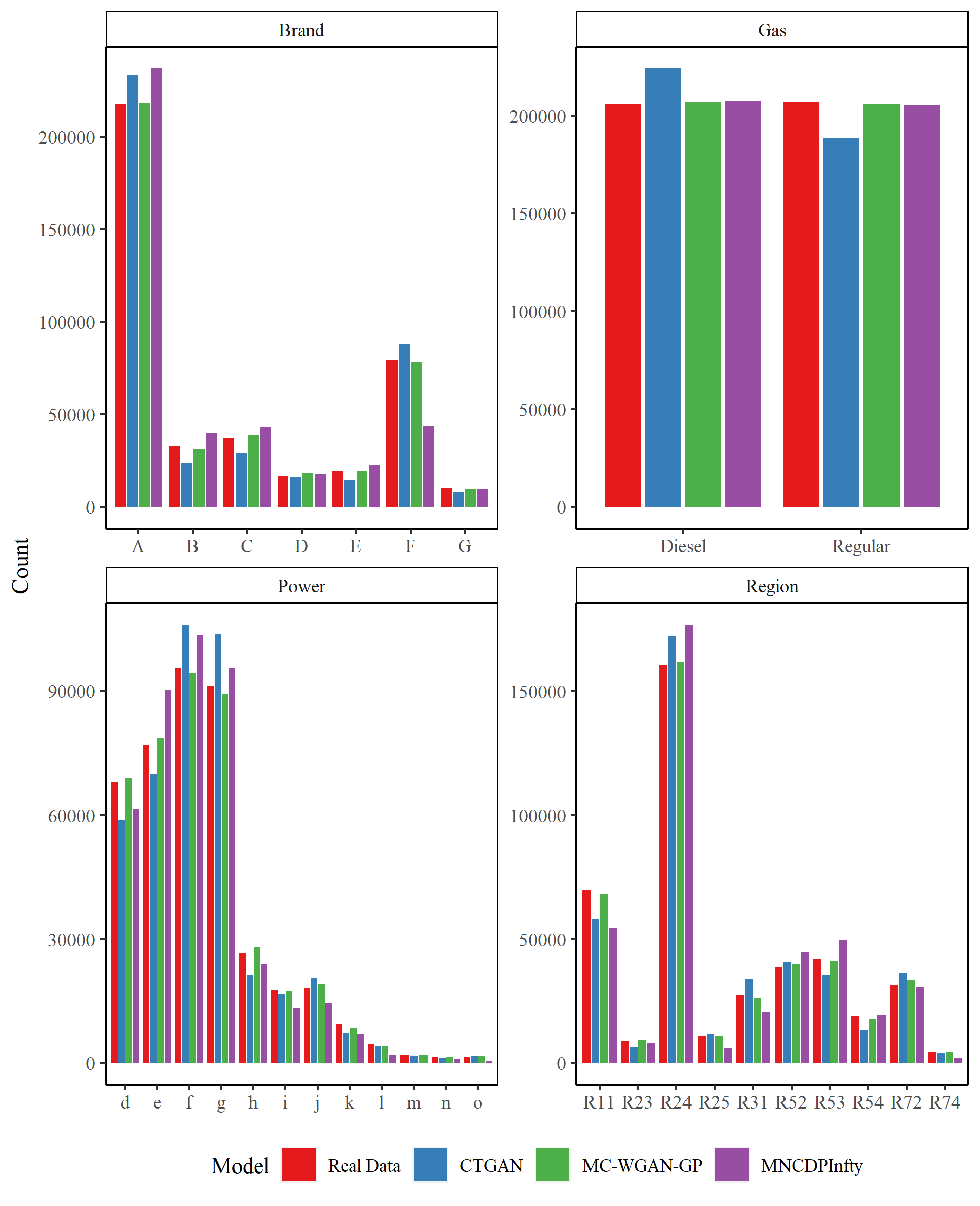}
    \caption{Comparison of univariate categorical variable distributions.}
    \label{fig:AllCatagoricalBars}
\end{figure}

We first compare the results for the categorical variables. Figure \ref{fig:AllCatagoricalBars} shows the observations in each category in the real and generated datasets for brand, gas type, power and region for the real data, the MC-WGAN-GP, the CTGAN and the MNCDP-GAN without DP. We see that from the univariate perspective, the three models all replicate the real data reasonably well. In particular, the MC-WGAN-GP (blue) reproduces closely the univariate distributions in the real data (red) for these four categorical variables. 

\begin{figure}[h]
\centering
    \includegraphics[width=\textwidth]{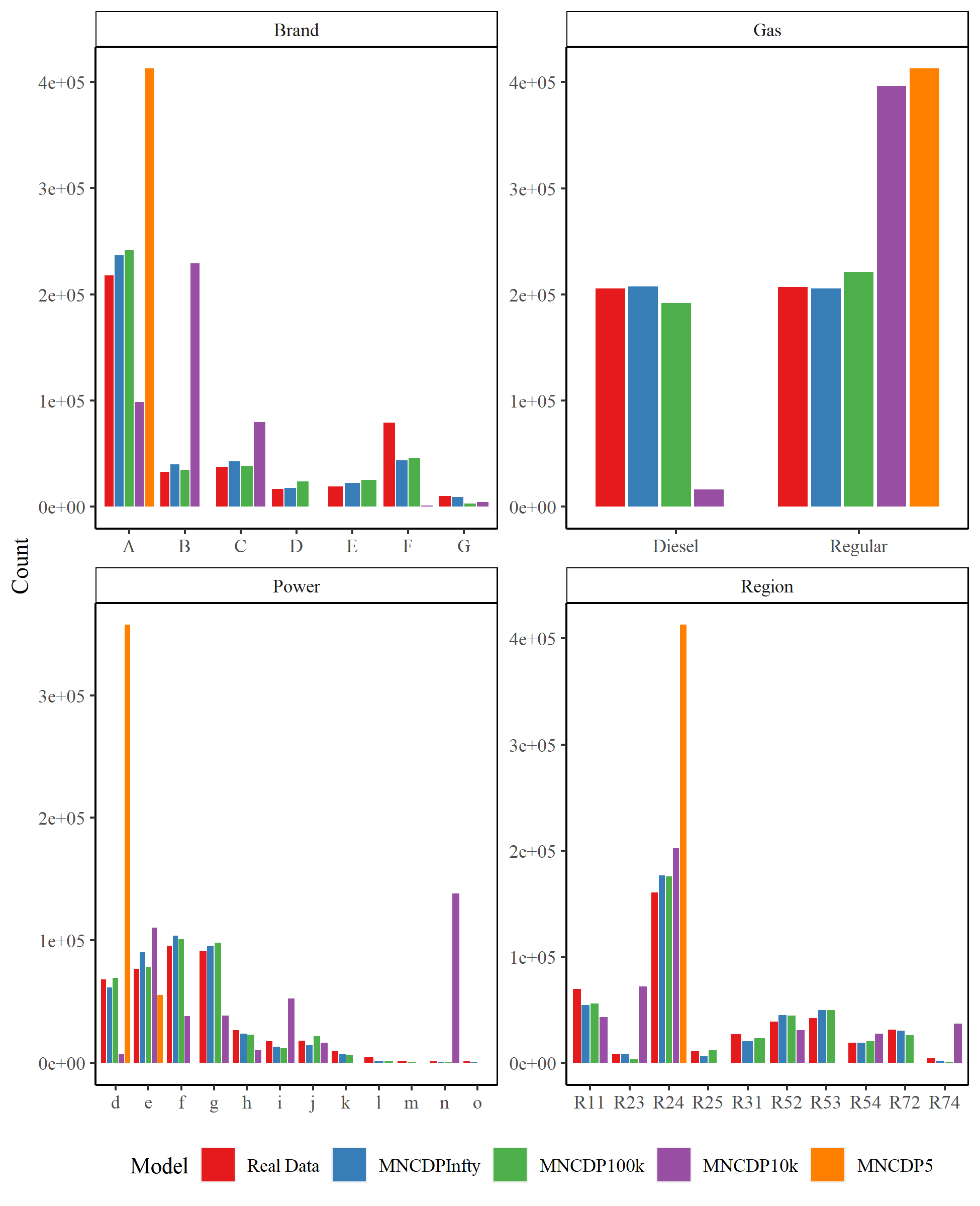}
    \caption{Comparison of univariate categorical variable distributions for MNCDP-GAN models with $\epsilon\in\{5, 10000, 100000, \infty\}$.}
    \label{fig:AllCatagoricalBars_DP}
\end{figure}

Figure \ref{fig:AllCatagoricalBars_DP} shows the same information as Figure~\ref{fig:AllCatagoricalBars} but for the MNCDP model with varying levels of differential privacy. It becomes readily apparent that the synthesized data gets dramatically worse as the level of differential privacy increases. Again, the model with $\epsilon = \infty$, no differential privacy, follows the data rather well. As $\epsilon$ decreases to 100,000, the model still maps relatively well to the real data. But, $\epsilon=10,000$ and especially $\epsilon = 5$, are not close to the real data. The noise added to the process in both cases completely obscures the original signal. This will be a consistent result throughout our case study.

\begin{figure}[h]
\centering
    \includegraphics[width=\textwidth]{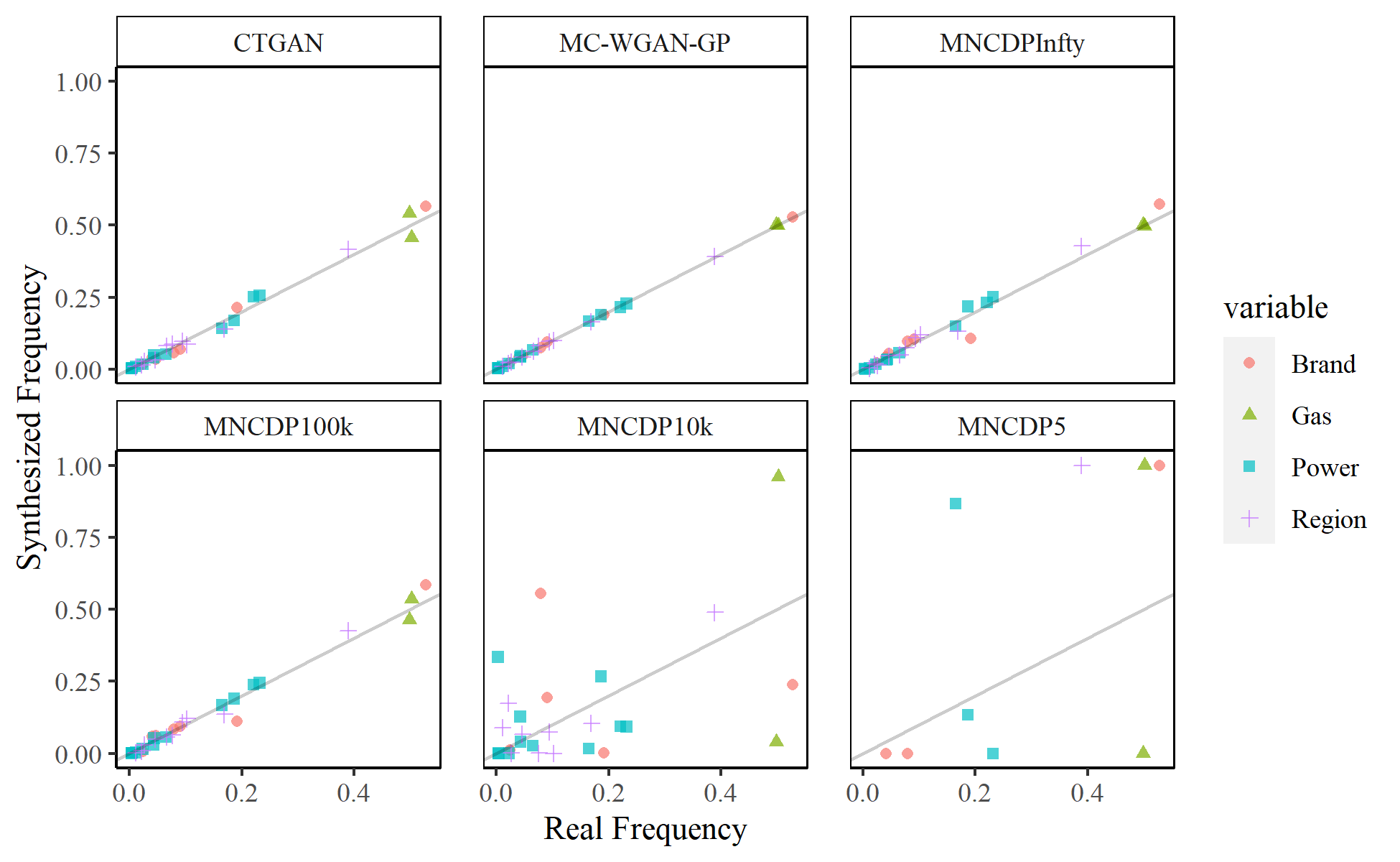}
    \caption{Comparison of the synthesized and real group frequencies for each class of the four categorical variables.}
    \label{fig:FreqScatter}
\end{figure}

Shown another way, Figure \ref{fig:FreqScatter} plots the frequency for each category in the real data on the $x$-axis against the frequency in the synthesized data on the $y$-axis for each of the GAN considered, color-coded by feature. The line $y=x$ is also plotted. The MC-WGAN-GP dataset seems to match the real frequencies the best, followed by the CTGAN, MNCDPInfty, and MNCDP100k models (which are relatively similar). As noticed above, the MNCDP10k and MNCDP5 models are drastically different from the original data. 

\begin{figure}[h]
\centering
    \includegraphics[width=\textwidth]{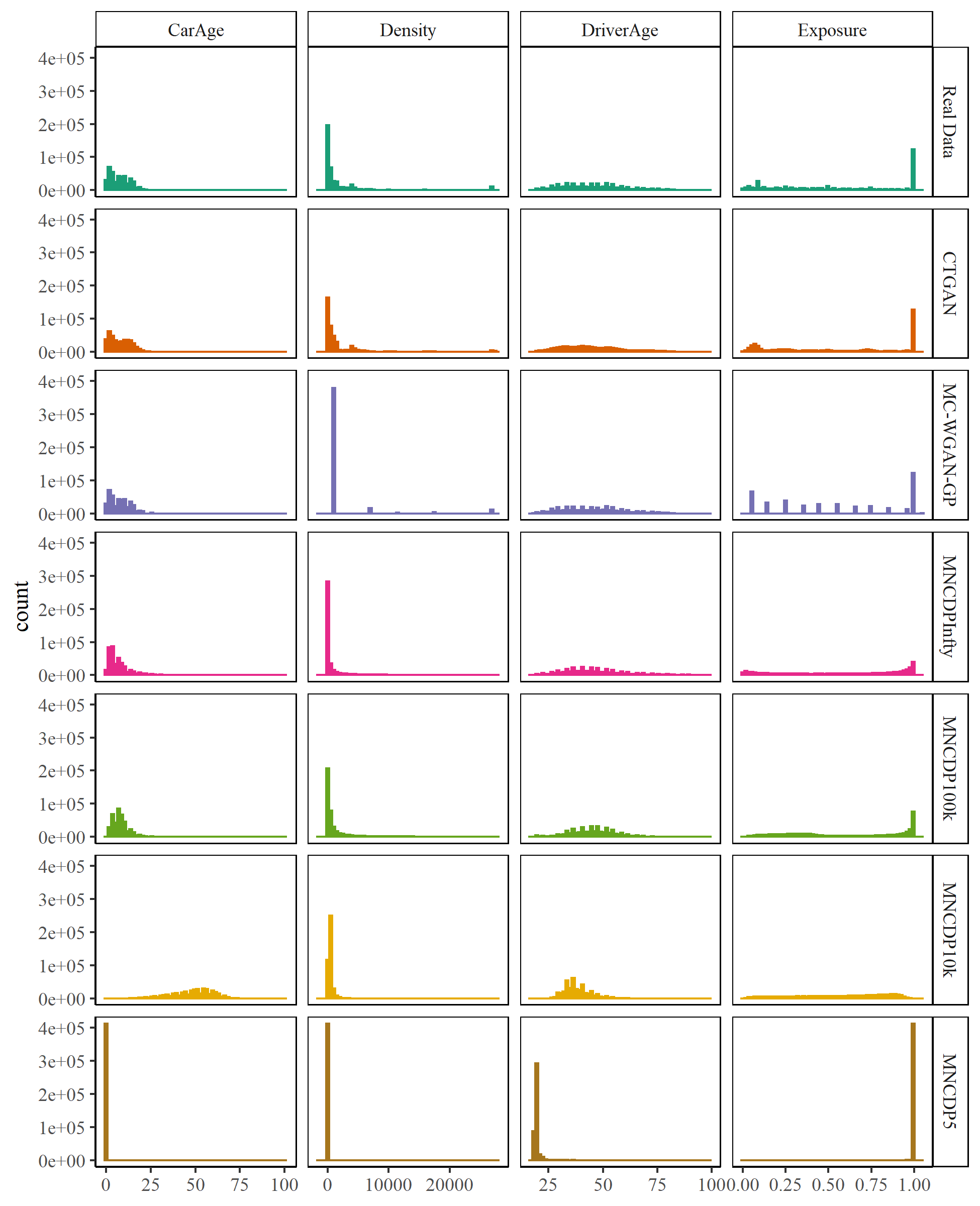}
    \caption{Comparison of univariate numerical variable distributions.}
    \label{fig:NumStacked}
\end{figure}

For the numeric variables CarAge, Density, DriverAge and Exposure, Figure \ref{fig:NumStacked} shows the distributions of the real data on the top row and compares it to the distributions of data generated by the models. One of the most difficult aspects of insurance data is the exposure variable. It has a large proportion which are exactly 1. After accounting for those, the rest of the data tends to be either close to 0 or close to monthly intervals (1/12, 2/12, 3/12, etc.). Both the CTGAN and the MC-WGAN-GP do well synthesizing the correct number of 1 values, but when looking at the rest of the distribution, the MC-WGAN-GP is too bumpy and the CTGAN might be too smooth. For the Density random variable, CTGAN is the best. Both DriverAge and CarAge are matched by the three methods rather well.  

\begin{figure}[h]
\centering
    \includegraphics[width=\textwidth]{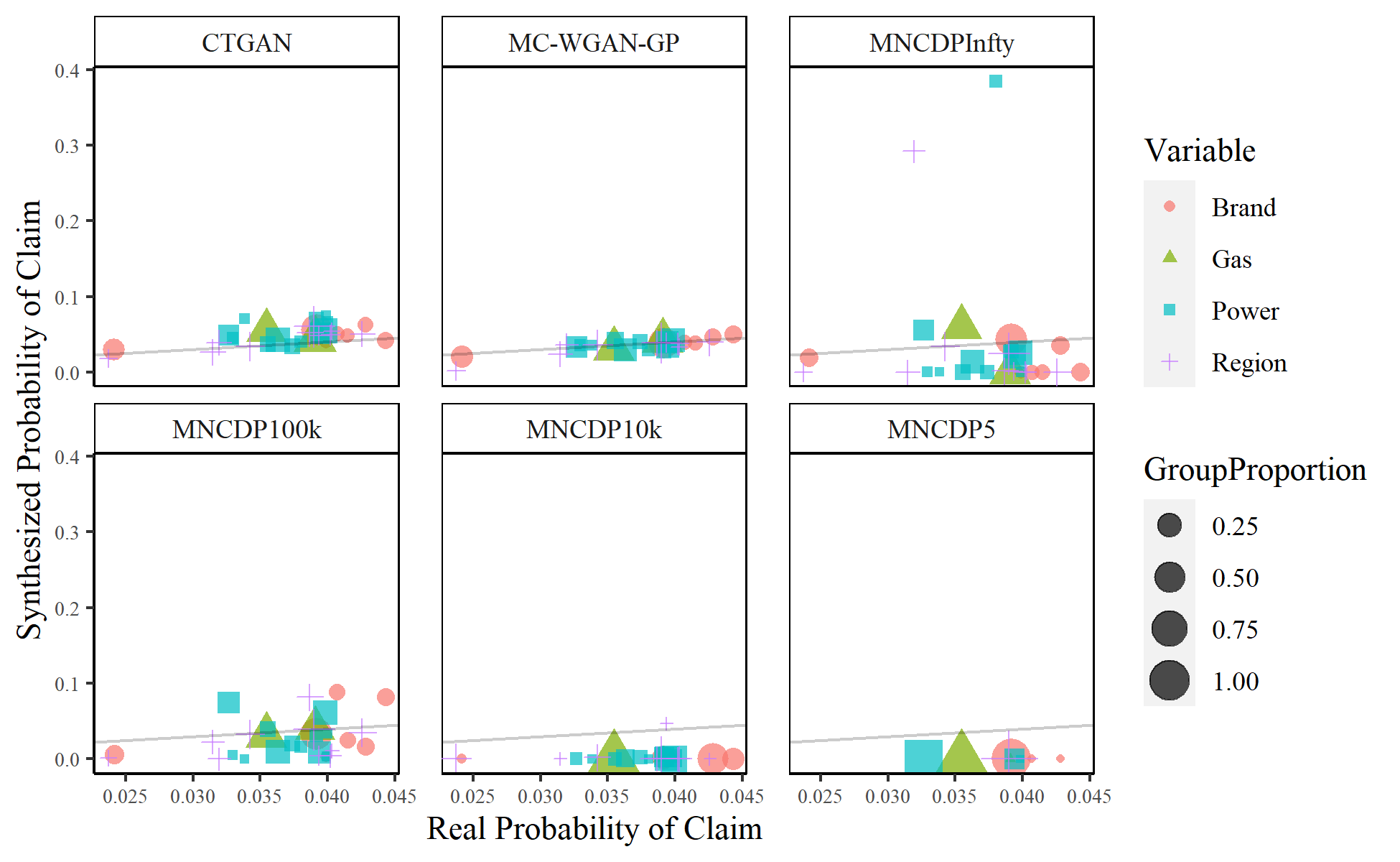}
    \caption{Comparison of the synthesized and real empirical probability of a claim for each group.}
    \label{fig:ProbScatter}
\end{figure}

It is important to correctly model the univariate characteristics, but it is even more important to correctly model the multivariate relationships. This is especially true with the relationship between claim counts and the various explanatory variables. Figure \ref{fig:ProbScatter} compares the probability of a claim in each categorical group. The real probability of a claim in on the $x$-axis while the synthesized probability is on the $y$-axis. The line $y=x$ is also plotted to show the ideal goal. Also, the size of the marks shows the proportion of the synthesized data in each group. By this metric, the MC-WGAN-GP performs the best, followed closely by the CTGAN. The MNCDP-GAN does not do well, even without differential privacy.

\begin{figure}[h]
\centering
    \includegraphics[width=\textwidth]{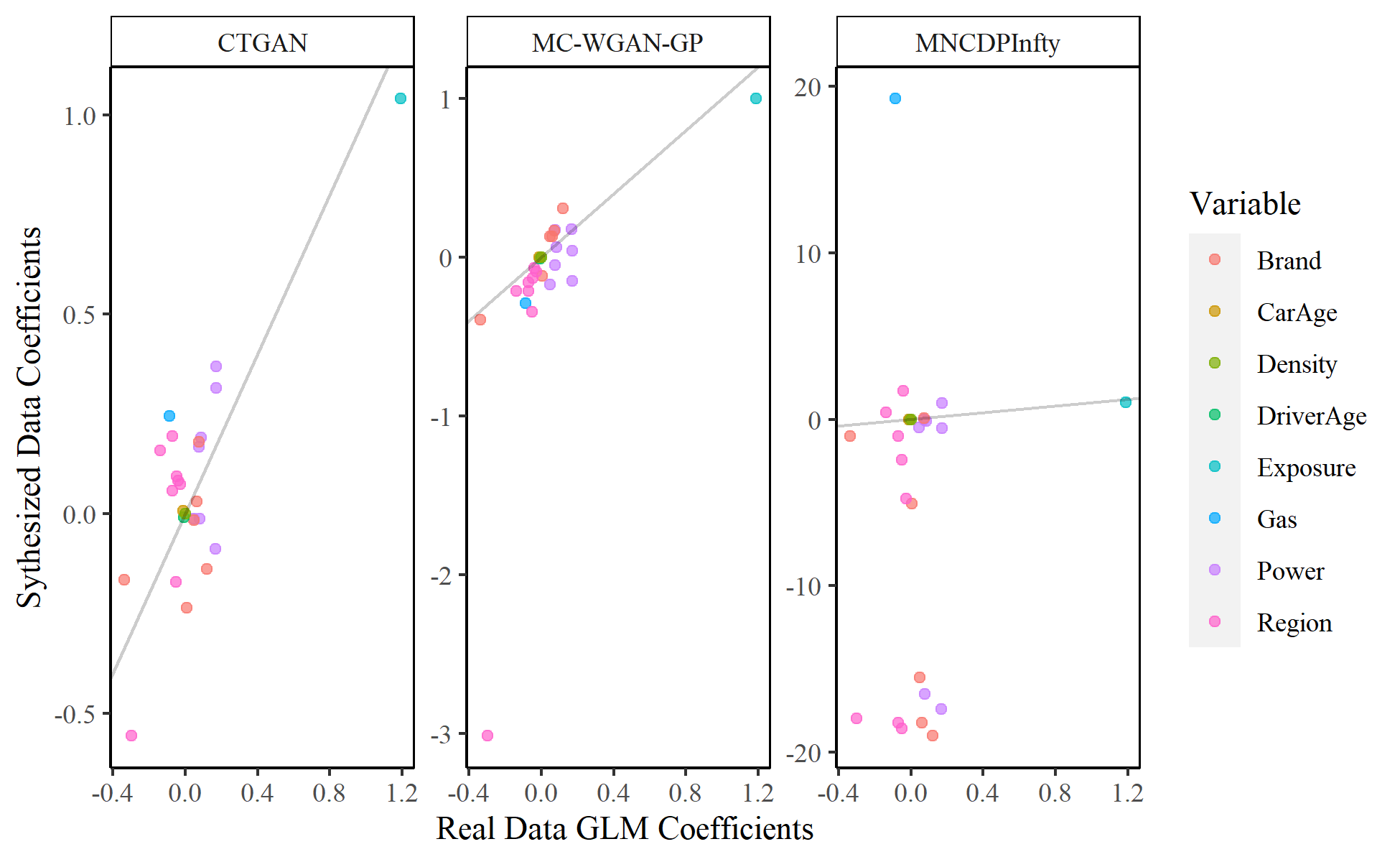}
    \caption{Comparison of the Poisson regression coefficients for the synthesized and real data.}
    \label{fig:GLMCoefficients}
\end{figure}

Our last test examines how consistent are models fitted on the original and the synthesized data. We split the real data into two parts, a 70\% training set and a 30\% test set. We fit a Poisson GLM on the training set predicting the number of claims using all eight of the explanatory variables. We then fit the same model on a 70\% sample from each of the synthesized datasets. The sampling and model fitting are performed 5,000 times to understand the sampling variability and to obtain more consistent estimates. In Figure \ref{fig:GLMCoefficients}, we compare the average estimated regression coefficients for each of the three models. We found that the CTGAN coefficients are all close to the real coefficients. The coefficients estimated with the MC-WGAN-GP data are similarly close with the exception of a single region coefficient. The results achieved using the MNCDP synthesized data are again the worst.

With each fitted model, we then predicted the claim counts for the 30\% real test data and compared the predictions from the models fit on the synthesized data to the predictions from the models fit on the real data. Table \ref{tab:RegErrors} shows the median absolute error (MAE) and mean squared error (MSE) between the predictions, with 95\% bootstrapped confidence intervals. The MC-WGAN-GP significantly outperforms the other two models on both metrics. The CTGAN performs next best and the worst performance is again attributed to the MNCDP-GAN.

\begin{table}[ht]
\centering
\caption{Poisson regression prediction errors (and bootstrapped 95\% intervals) for the three main GANs.} 
\label{tab:RegErrors}
\begin{tabular}{lrr}
  \hline
Model & MAE ($\times1000$)~~ & MSE ($\times1000$)~ \\ 
  \hline
MC-WGAN-GP & $5.0~(~4.68, ~~5.44)$ & $0.08~(0.06, 0.10)$ \\ 
  CTGAN & $10.8~(10.28, 11.32)$ & $0.38~(0.36, 0.40)$ \\ 
  MNCDPInfty & $32.8~(32.38, 33.22)$ & $3.36~(3.20, 3.52)$ \\ 
   \hline
\end{tabular}
\end{table}

\section{Conclusion}
\label{sec:conclu}

In this paper, we presented, implemented and compared three different methods to synthesize insurance data. All of the methods are based on generative adversarial networks. All three methods have advantages and disadvantages. The MC-WGAN-GP method synthesized the most realistic data. It generated data which was very similar (accounting for both univariate and multivariate relationships) to the real data. The CTGAN method was the easiest to use, especially for someone more familiar with \textsf{R} than Python. The CTGAN synthesized data was almost as good as the MC-WGAN-GP data. The main drawbacks of MC-WGAN-GP and CTGAN is that there are no privacy guarantees, meaning that some records in the generated data could still contain confidential information. The MNCDP-GAN incorporates differential privacy, but the synthesized data (even without differential privacy) was not as good as with the other two methods.

Future work can start from any of the three models and attempt to add the advantages of the other two. Either add differential privacy and ease of use to the MC-WGAN-GP, improved synthesis and differential privacy to the CTGAN, or improved synthesis and ease of use to the MNCDP-GAN. In any case, GANs are a promising tool for synthesizing and protecting private and important data. 

\section{Acknowledgments}

This work was supported by a Casualty Actuarial Society (CAS) Individual Grant and by M.-P. C\^ot\'e's Chair in Educational Leadership in Big Data Analytics for Actuarial Sciences --- Intact. The authors thank the members of the CAS project oversight committee, Syed Danish Ali, Morgan Bugbee, Marco De Virgilis, and Greg Frankowiak, for useful feedback throughout the project. The project would not have been possible without the support provided by Calcul Qu\'ebec (www.calculquebec.ca), Compute Canada (www.computecanada.ca), and the statistics department computing cluster at BYU.

\section*{References}

\bibliographystyle{elsarticle-harv}
\bibliography{GANCitations}

\clearpage 
\clearpage

\appendix
\section{MC-WGAN-GP Hyperparameter Tuning}
\label{app:B}

\noindent In the MC-WGAN-GP model, we tuned the following hyperparameters.
\begin{itemize}
    \item Loss Penalty -- loss\_penalty
    \item Generator Batch Norm Decay -- gen\_bn\_decay
    \item Discriminator Batch Norm Decay -- disc\_bn\_decay
    \item Generator L2 Regularization -- gen\_L2\_reg
    \item Discriminator L2 Regularization -- disc\_L2\_reg
    \item Learning Rate -- learning\_rate
\end{itemize}

We used a random search to explore the settings and decided on the values in Table~\ref{tab:hyperSettings}.

\begin{table}[h]
    \centering
        \caption{Hyperparameter settings for the MC-WGAN-GP.}
    \label{tab:hyperSettings}
    \begin{tabular}{lll} \hline
        Hyperparameter & Explored Values & Chosen Value\\ \hline
        loss\_penalty & 1, 5, 10, 20, 50 & ~10 \\
        gen\_bn\_decay & 0, 0.10, 0.25, 0.45, 0.50, 0.90 & ~0.90 \\
        gen\_L2\_reg & 0, 0.00001, 0.0001, 0.001, 0.01 & ~0 \\
        disc\_L2\_reg & 0, 0.00001, 0.0001, 0.001, 0.01 & ~0 \\
        learning\_rate & 0.001, 0.005, 0.01 & ~0.01\\ \hline
    \end{tabular}
\end{table}

\section{MNCDP-GAN Methodology}
\label{app:A}

In this appendix, we give details on the preprocessing of the variables in the French Motor Third-Party Liability frequency dataset for the MNCDP-GAN in~\ref{app:prep}. We then explain the training procedure in~\ref{app:train}, the hyperparameter optimization in~\ref{app:hyp} and the final setting in~\ref{app:fin}.

\subsection{Preprocessing}
\label{app:prep}



Because the Exposure variable should be capped at one, all 421 samples whose Exposure was above this threshold were removed. As for the target ClaimNb, it was converted to a categorical variable since there are only five possible values (0 to 4) and it is highly skewed towards zero. The adjusted dataset, which was the one used to train the models, contains 412,748 samples explained by four numerical (DriverAge, CarAge, Exposure and Density) and five categorical (ClaimNb, Power, Brand, Gas and Region) variables. 

For the MNCDP-GAN experiments, multiple configurations differing on the types of those variables were tested. The first one, called "baseline", uses this data as is. In the "all-cat" configuration, both Exposure and Density were binned into categorical variables, as was also done for the MC-WGAN-GP. In the "bin" configuration, the four numeric variables were binned in order to be treated as categorical. Expert insight was required to determine the binning size for each feature independently. 

In the case of DriverAge, ten bins of increasing size (more bins at younger ages) were made to approximate a normal distribution. For CarAge, a single bin was used for the value 0 while the rest was split using ten quantiles. The resulting distribution is thus closer to the uniform distribution. For Exposure, 12 bins were chosen to reflect the 12 months of a year with the exception of the value one which has a bin of its own. The resulting distribution is skewed towards that last bin. Finally, for Density, the binning was applied on its natural logarithm and based on the deciles, so that the resulting distribution is almost uniform. 

\subsection{Training}
\label{app:train}


For each configuration of the MNCDP-GAN, the autoencoder (AE) and the GAN were trained independently one after the other, since the GAN requires the decoder. The training of the AE was done over 20,000 iterations, which was determined to be sufficient for convergence. Before feeding the preprocessed data to the network, the numerical features were normalized using min/max normalization while the categorical variables were encoded using one-hot vectors. Unlike the more common way to encode $N$ categories in $N-1$ dimensions, it was decided to use one dimension per category instead. Otherwise, the GAN would never generate the category not having a dimension of its own. The AE uses the binary cross entropy loss.

The training of the GAN was done over 2M iterations. This value was determined empirically based on the results obtained. However, because of the known difficulty to train GANs (no stability guarantees), it could be adjusted depending on the configuration and the hyperparameters. The generator and discriminator use the zero-sum objective function as proposed by \cite{Arjovsky/etal:2017} for their learning. As recommended in this same paper, the discriminator was updated more often than the generator in order to train until optimality. Using a linear activation as the output layer of the discriminator and the RMSProp optimizer for the GAN are also recommendations from these authors that were applied.

For both the AE and the GAN, the preprocessed dataset was split 2/3 for training and 1/3 for validation. The basic architecture of the networks (the number and sequence of layers) was not changed from \cite{Tantipongpipat/etal:2019}. Unless stated otherwise, the activation functions used were always LeakyReLU with a negative slope of $0.2$. Because they depend on the input size and some hyperparameters (such as the latent dimensions), the sizes of the layers vary from configuration to configuration. The design choices of using the ADAM optimizer with gradient penalty for the AE, choosing the absolute bounds to clip the values of the WGAN gradients and choosing layer normalization over batch normalization for the generator follow the recommendations of \cite{Gulrajani/etal:2017}. 

The algorithm of the differentially private stochastic gradient descent (DP-SGD) can be found in \cite{Tantipongpipat/etal:2019}. For the differential privacy aspect, computing the L2 clipping norms of the gradients for the decoder and the discriminator was done as recommended by \cite{Abadi/etal:2016}. Finally, before training the models used to obtain results, the hyperparameters were tuned using a random search.

\subsection{Hyperparameter optimization}
\label{app:hyp}

As for the training, the tuning of the hyperparameters of the AE and the GAN was done in two stages. In a random search, each combination of hyperparameters tested is selected randomly from the chosen grid. The number of search iterations is set based on a time/resources compromise. The strategy was to run multiple searches instead of a single big one. Each time, the search space was narrowed for more fine-tuning. In all cases, the tuning was done in a non private way ($\epsilon=\infty$) in order to reduce computation time.

For the AE, the hyperparameters tested were the minibatch size, compression dimension, learning rate, $\beta_1$ and $\beta_2$ parameters of the ADAM optimizer, and the L2 penalty of the weight decay for the optimizer. In first experiments, these last three hyperparameters did not affect the training significantly so it was decided to leave them at their usual default values (0.9, 0.999 and 0, respectively). To evaluate the performance of each combination, the validation and training losses were saved and sorted. The combination with the lowest final validation loss was chosen. A regular training was then done to confirm that it was not overfitting.

For the GAN, the hyperparameters tested were the minibatch size, the latent dimension of the generator, the learning rate, the number of iterations of the discriminator before updating the generator once, the L2 penalty of the weight decay of the optimizer and the $\alpha$ smoothing constant of the RMSProp optimizer. Once again, these last two hyperparameters of the optimizer did not affect the results significantly so they were left at 0 and 0.99 respectively. The evaluation of the performance of the GAN was not as straightforward as for the AE. Both the losses of the discriminator and the generator were plotted. Over time, it was found that the desired loss curve of the discriminator was one which dropped rapidly near $0$ and that then converged to that value over training iterations. For the generator, a loss oscillating rather slowly around 0 (going positive for many thousands of iterations then going negative and so on) appeared a good indicator of performance. Fortunately, these tendencies could be spotted for training of only a few tens of thousands of iterations. Hence, to reduce computation time, the number of iterations for each combination was limited to 100,000.

Once a couple of potentially good combinations were identified in this manner, a full training over 2M iterations was done for each one. The generated samples of these trained models were then evaluated in respect to the univariate distributions of each variable (vs the real distributions). The predictions on the target ClaimNb of a random forest regressor and of a random forest classifier were also compared between the generated and the real samples. The combination giving the best overall results was kept. Because of the lack of stability of the GANs (even for the same hyperparameters and configuration), this best combination of hyperparameters was trained at least two more times with different seeds for the random number generator. The model of whichever run gave the best results was saved and used for the final results.

For both the autoencoder and the GAN, for a given configuration, the hyperparameters that had the most impact on the results were the minibatch size and the learning rate. When training differentially private models, the values of the hyperparameters were the same as those of its corresponding non private configuration. To guarantee the privacy, both the autoencoder and the GAN were trained from scratch (i.e. their non private counterpart were not used at any point).

\subsection{Configurations}
\label{app:fin}


As stated previously, different configurations were tested to see the impact of changing the types of some features. These configurations as well as the values of their hyperparameters are listed in Table~\ref{tab:params}. Note that, except for the types of the features, the "baseline" and "all\_cat" configurations share the same hyperparameters. This is because using the values of the first on the second gave good results. 

\begin{table}[ht]
\centering
\caption{\label{tab:params} Hyperparameter values for the different configurations.}
	\begin{tabular}{lllll}
	\hline
		\multicolumn{2}{c}{Hyperparameters} & \multicolumn{3}{c}{Configuration} \\ 
		&  & baseline      & all\_cat      & bin     \\ \hline
		\multirow{4}{*}{Features} & DriverAge & Numerical      & Numerical      & Categorical     \\
		& CarAge & Numerical      & Numerical      & Categorical     \\
		& Density & Numerical      & Categorical      & Categorical     \\
		& Exposure & Numerical      & Categorical      & Categorical     \\ \hline
		\multirow{7}{*}{AE} & l2 norm clip & 0.022      & 0.022      & 0.022     \\
		& Minibatch size & 64      & 64      & 128     \\
		& Compression dim & 25      & 25      & 50     \\
		& Learning rate & 0.01      & 0.01      & 0.01     \\
		& $\beta_1$ (ADAM) & 0.9      & 0.9      & 0.9     \\
		& $\beta_2$ (ADAM) & 0.999      & 0.999      & 0.999     \\
		& L2 penalty & 0      & 0      & 0     \\ \hline
		\multirow{8}{*}{GAN} & l2 norm clip & 0.027      & 0.027      & 0.027     \\
		& Clip value & 0.01      & 0.01      & 0.01     \\
		& Minibatch size & 128      & 128      & 128     \\
		& Latent dim & 25      & 25      & 30     \\
		& Learning rate & $4.5\times10^{-5}$      & $4.5\times10^{-5}$      & $3.9\times10^{-5}$     \\
		& Discriminator updates & 10      & 10      & 5     \\
		& Alpha (RMSProp) & 0.99      & 0.99      & 0.99     \\
		& l2 penalty & 0      & 0      & 0    
	\end{tabular}
\end{table}

During the training of the autoencoder, the learning rate was reduced by a factor 0.2 when the validation loss reached a plateau (tolerance of $1\times10^{-4}$) for 1000 iterations (i.e. the patience). Its minimum value was limited to $1/100$ of the initial learning rate shown in Table~\ref{tab:params}. Two other hyperparameters choices were tested, but are not listed in Table \ref{tab:params} as they did not improve the results. The first was using a Kaiming uniform initialization of the weights for the AE and GAN instead of the default pytorch initialization. The second was using the ADAM optimizer for the GAN instead of RMSProp. Both optimizers gave almost the same results when using the same random seed.

\end{document}